\pdfoutput=1
\PassOptionsToPackage{table}{xcolor}
\documentclass[11pt]{article}

\usepackage{EMNLP2023}

\usepackage{times}
\usepackage{latexsym}
\usepackage{booktabs}

\usepackage{multirow}
\usepackage{stfloats}
\usepackage{caption}

\usepackage[T1]{fontenc}

\usepackage[utf8]{inputenc}

\usepackage{microtype}

\usepackage{inconsolata}

\usepackage{graphicx}
\usepackage[colorinlistoftodos]{todonotes}
\usepackage{diagbox}
\usepackage[table]{xcolor}

\title{Acquiring Linguistic Knowledge from Multimodal Input}

\author{Theodor Amariucai \\
  ETH Zürich \\
  \texttt{tamariucai@ethz.ch} \\\And
  Alex Warstadt \\
  ETH Zürich \\
  \texttt{awarstadt@ethz.ch} }

\begin{document}
\maketitle
\begin{abstract}
In contrast to children, language models (LMs) exhibit considerably inferior data efficiency when acquiring language. 
In this submission to the BabyLM Challenge \citep{warstadt2023findings}, we test the hypothesis that this data efficiency gap is partly caused by a lack of multimodal input and grounding in the learning environment of typical language models.
Although previous work looking into this question found that multimodal training can even harm language-only performance, we speculate that these findings can be attributed to catastrophic forgetting of complex language due to fine-tuning on captions data. 
To test our hypothesis, we perform an ablation study on FLAVA \citep{flava}, a multimodal vision-and-language model, independently varying the volume of text and vision input to quantify how much text data (if any) can be offset by vision at different data scales.
We aim to limit catastrophic forgetting through a multitask pretraining regime that includes unimodal text-only tasks and data sampled from WiT, the relatively diverse Wikipedia-based dataset \citep{srinivasan2021wit}.
Our results are largely negative:
Multimodal pretraining does not harm our models' language performance but does not consistently help either.
That said, our conclusions are limited by our having been able to conduct only a small number of runs.
While we must leave open the possibility that multimodal input explains some of the gap in data efficiency between LMs and humans, positive evidence for this hypothesis will require better architectures and techniques for multimodal training.
\end{abstract}

\section{Introduction}
Children can learn language from a relatively small amount of linguistic input: at most 100 million words \citep{gilkerson2017mapping}. 
By contrast, the quantity of training data a language model needs to achieve strong grammar and language performance is on the order of billions or tens of billions of words \citep{zhang2020you}. 
This data efficiency gap may be due, in part, to innate differences in learning mechanisms between models and humans, but environmental differences likely play a role as well \citep{warstadt2022what}.
This work tests the hypothesis that the lack of visual grounding in language models accounts for some of the gap in data efficiency.

The likelihood of finding evidence for this hypothesis rests largely on two factors: (1) its cognitive plausibility and (2) its technological viability. If vision does help children learn language, then there ought to be some way of incorporating vision into text-only language models that improves their learning ability. However, whether or not we can find this approach depends on the present technological capabilities of multimodal models.

To address the first point, one cognitively-motivated mechanism for how children integrate nonlinguistic sensory data in language learning is cross-situational learning (XSL) \citep{xsl_book}.
This theoretical mechanism holds that the learner accumulates statistical evidence about word meanings by observing multiple instances of co-occurring word-object pairs across many different real-world situations \citep{smith2011cross,xsl_is_implicit_and_strategic, xsl_behavioral_and_computational}.
Encouragingly, \citet{xsl_neuralnets} find that, in a highly constrained visual-linguistic domain, computational multimodal models \emph{do} benefit from cross-situational learning.

To address the second point, prior evidence that vision will improve language models given current technologies is, at best, mixed.
Recent approaches have successfully trained Transformer-based multimodal language models using self-supervised objectives resembling those developed originally for the training of unimodal models \citep{tan2019lxmert}.
Nevertheless, in comparison to the unimodal models, multimodal LMs often perform relatively poorly on language-only tasks \citep{iki_and_aizawa_2021}.
We hypothesize that these shortcomings may be due to the common practice of training multimodal models by fine-tuning pretrained language models on captions data.

Our approach addresses these limitations in two ways that differ from most previous pretraining recipes. 
First, we use the FLAVA architecture and follow its multitask training procedure \citep{flava}. 
Second, we train on the Wikipedia-based WiT dataset \citep{srinivasan2021wit}, which pairs images with a mixture of strongly aligned (but formulaic) captions and weakly aligned (but syntactically complex) articles.
Despite these efforts, our results show that the addition of image data and multimodal training objectives leads to no reliable improvement over text-only baselines on benchmarks for grammar \citep[BLiMP;][]{warstadt2020blimp}, understanding \citep[GLUE;][]{wang2019glue}, and generalization \citep[MSGS;][]{warstadt2020learning}. 
We conclude that, to the extent that multimodality is partly responsible for the data-efficiency gap, present multimodal (and multitask) pretraining methods do not benefit from this richer learning signal.

To summarize, this work brings forward three main contributions:
\begin{enumerate}
    \item We develop a robust codebase\footnote{\href{https://github.com/amariucaitheodor/acquiring-linguistic-knowledge}{https://github.com/amariucaitheodor/acquiring-linguistic-knowledge}} for pretraining (from scratch) large multimodal LMs under varying text and vision input configurations.
    \item We evaluate, in this controlled environment, the effects of the visual signal on the model's textual encoder (hence, its linguistic ability).
    \item We investigate plausible mechanisms for how incorporating visual input into the pretraining procedure might affect linguistic behavior.
\end{enumerate}

\section{Background}\label{sec:background}

Prior work on multimodal language model training can be roughly differentiated by whether the main objectives are cognitively-oriented or engineering-oriented. So far, neither of these directions has produced clear evidence supporting the hypothesis that multimodality aids language learning at the scale of human language acquisition. Many cognitively oriented contributions are limited by a small data scale or a restricted domain. By contrast, engineering-oriented contributions using state-of-the-art Transformer-based architectures achieve more developmentally plausible scale and diversity but emphasize multimodal performance over language learning.

\subsection{Cognitively Oriented Approaches}

Infants enter a diverse and abundant visual world where they develop mental models to comprehend and mimic the patterns they encounter. These mental models empower them to grasp and anticipate their surroundings and accomplish objectives by incrementally improving their communicative abilities \citep{cell}. Relatedly, the impact of vision on specific aspects of human language learning has been an important question in human development for decades.

Contemporary research has tried to answer this question through computational simulations of cognitive processes involved in language acquisition. 
Multimodal models trained on visual question answering or reference games can use cross-situational learning to learn grounded meanings of words \citep{mao2019neurosymbolic,wang2021languagemediated,xsl_neuralnets,portelance2023learning}.
Nonetheless, computational models show different learning biases than humans in many cases, at least in the absence of specific training or architectural interventions \citep{gauthier2018word,vong2022cross}.
Ultimately, however, all of these studies are limited in their cognitive plausibility and language learning by a reliance on supervised training on small, artificial datasets in which texts and images correspond to arrangements of a limited set of objects in a simple, usually static scene.

Other studies aim for more naturalistic training.
\citet{lazaridou2017multimodal} and \citet{chrupala2015learning} are notable for pioneering self-supervised training objectives for multimodal models several years before the advent of Transformer architectures trained on masking objectives.
\citet{wang2023finding} train LMs on data from the SAYCam dataset \citep{saycam}, pairing (written) child-directed utterances with visual data from the child’s point of view.
While this data domain is nearly ideal from a developmental plausibility perspective, the available data is too small to model anything past the first month of development.

Finally, we note that most of the studies in this area focus primarily on word learning.
However, the data efficiency gap applies more broadly to language learning.
Recent studies evaluating contemporary Transformer-based models have largely reported negative results for the effect of multimodality on semantics \citep{shahmohammadi2022language}, commonsense reasoning \citep{yun2021doesa}, and learning biases \citep{kuribayashi2023does}. 
To the best of our knowledge, ours is the first work to perform targeted syntactic evaluation \citep{marvin2018targeted,warstadt2020blimp,hu2020systematic} on multimodal models.

\subsection{Engineering Oriented Approaches}\label{subsec:engineering_approaches}

The most effective recent approaches for training multimodal language models generalize the self-supervised objectives (and Transformer-based architectures) that have become dominant for unimodal language models such as BERT \citep{bert} to the vision-and-language domain \citep{li2019unicodervl,li2019visualbert,zhou2019unified_vlp,tan2019lxmert,chen2020uniter,yu2021ernievil, ViLBERT,vision_unmasked, pramanick2023volta}. 

These studies, for the most part, share many aspects of a typical recipe: 
First, they initialize all or some of the model parameters with the pretrained weights of a model such as BERT. 
Second, they fine-tune (using one or more self-supervised objectives) on a dataset of image-caption pairs.%
\footnote{
For example, in the masked multimodal modeling task \citep[MMM;][]{tan2019lxmert}, regions of an aligned image-text pair are randomly masked before being input into the model and then predicted.
As the information from the image presumably helps reconstruct the masked text \citep{vision_for_language}, this objective encourages learning text representations that encode information from the visual modality (and vice-versa).
}
Finally, the model is evaluated on multimodal tasks such as visual question answering or image captioning.

While the ability to perform such grounded tasks is the key advantage of multimodal models over unimodal ones, it is critical for our research question to examine whether this advantage comes at the cost of language ability. 
Unfortunately, few of the works that train new multimodal models evaluate on language-only tasks. 
Some works perform this evaluation post hoc.
\citet{iki_and_aizawa_2021} study five multimodal architectures, all initialized with BERT and fine-tuned using identical data and training objectives by \citet{vision_unmasked}. 
Evaluating on the GLUE benchmark \citep{wang2019glue}, they find that, in nearly all cases, the original pretrained BERT outperforms the models with additional multimodal fine-tuning. 
Similar results are reported by \citet{madasu2023multimodal} and \citet{yun2021doesa}.

From a human development perspective, it may seem unintuitive that additional supervision on images harms language performance. 
However, from a machine learning perspective, this finding is easy to explain as an example of domain mismatch \citep{yun2021doesa}, catastrophic forgetting \citep{mccloskey1989catastrophic}, under-parameterization \citep{msc_thesis}, or other similar technical reasons.

BERT's original training data (Wikipedia and books) is diverse in terms of writing style and subject matter. 
By contrast, captions datasets commonly used to train multimodal LMs, such as MS COCO \citep{chen2015microsoft} or Visual Genome \citep{krishna2017visual}, consist entirely of short formulaic physical descriptions of objects or scenes. 
Hence, the text domain that the models were trained on most recently bears little resemblance to the texts in the GLUE tasks, for example.
Furthermore, the multimodal tasks incentivize using the models' limited parameters for both text \emph{and} image processing, potentially sacrificing language ability. 

Our experiments, which we describe in Section \ref{sec:methods}, are designed to address these issues through two complementary approaches: 
First, we prevent catastrophic forgetting by multitask-training on the language-only masked language modeling (MLM) objective jointly with the multimodal objectives. 
Second, we lessen the impact of domain mismatch by training on data that pairs images, not just with captions, but also with longer and more complex texts.

\section{Methods}\label{sec:methods}

We conduct experiments to uncover differences in how language models' linguistic abilities change as the amount of visual input varies. 
We pretrain and evaluate multimodal LMs in eight conditions, derived by independently varying the volume of text data (10M or 100M words) and image data (none, 40K, 400K, or 4M images). 
We perform only one training run for each of the eight conditions due to computing constraints (see Limitations, Section \ref{sec:limitations}).
The text quantities are compatible with both human-scale linguistic exposure \citep{gilkerson2017mapping} and the BabyLM strict-small and strict tracks \citep{warstadt2023findings}.

\subsection{Dataset} \label{subsec:dataset}

All the data for our experiments comes from WiT, a large, multimodal dataset entirely sourced from Wikipedia \citep{srinivasan2021wit}. 
Our choice of WiT was motivated by its size and the diversity and complexity of its text data. 
English WiT includes 5.5M image--text pairs,\footnote{WiT is also multilingual, containing over 30M pairs in over 100 languages. During preprocessing, however, we only sample English text.} making it one of the largest public datasets of its kind.
It contains extended passages from Wikipedia articles, offering a more representative sample of sentence types than typical multimodal datasets sourced from captions.
Furthermore, WiT features multiple types of text aligned with a given image.
From most strongly aligned to most weakly aligned, these include alt text, captions, article text from the same section as the image, and article text from the lead section.
Together with the fact that Wikipedia covers many different concepts and real-world entities, we hypothesize that WiT provides an adequately rich environment for supporting cross-situational learning while maintaining strong grammar and language understanding performance.

We subsample from the English portion of WiT to reach the desired data volume for each modality. For training purposes, we use either one (when either modality is 0\%) or three (when both modalities are non-zero) data loaders. For example, when training on 100M words and 40K images, we sample the first 10\% of the pairs for the text unimodal data loader, the first 1\% for the vision unimodal data loader, and the first $1\%=\min(10\%,1\%)$ for the multimodal data loader (containing paired images and text). Hence, all images in this configuration will be paired with some text, but not all texts will be paired with an image. This logic also implies that some images and texts will be seen both in the multimodal and their corresponding unimodal data loaders.

\subsection{Model}\label{subsec:model}
For our experiments, we use the FLAVA model architecture and training objectives \citep{flava}.
We choose to study FLAVA for two reasons:
First, \citet{flava} conduct a controlled comparison between a unimodally trained FLAVA text encoder and a fully multimodal FLAVA, and they report improved performance on language-only tasks from the multimodal model.
As such, FLAVA is the only example of a large multimodal model for which prior (anecdotal) evidence supports our hypothesis that vision can help language learning.
Second, FLAVA is trained in a multitask setting on a combination of unimodal text, unimodal vision, and multimodal objectives.
This methodology addresses our concern (Section \ref{subsec:engineering_approaches}) that other common multimodal training recipes can lead to catastrophic forgetting of linguistic ability.

FLAVA's architecture combines three modality-specific encoders: 
Text and vision embeddings are fed into unimodal text and vision encoders, respectively, and the hidden states output by these encoders are concatenated before being fed into a multimodal encoder.
For the unimodal objectives, task-specific heads can be placed after the corresponding unimodal encoder. All encoders are based on the ViT-B/16 encoder \citep{vit_image_transf}.%
\footnote{\emph{B/16} refers to a base-sized architecture with 86M total parameters using a patch resolution of 16x16. We opt for this version of the encoder based on the authors' observation that ViT-B/16 performs just as well as the larger alternatives when pretraining on smaller datasets of under 300M images (such as WiT).}


Following the original work, we pretrain models from scratch using multitask learning with the following five objectives: masked image modeling, masked language modeling, masked multimodal modeling for both text and vision, image-text matching, and cross-modal contrastive learning. More details on each objective, as well as the encoder architecture itself, can be found in the original paper \citep{flava}.

\subsection{Training Details}\label{subsec:training_config}

\paragraph{Hyperparameters}
We perform a hyperparameter search and empirically settle on the following values: 

\begin{itemize}
\itemsep-0.4em 
    \item Warmup steps: $10^4$
    \item Batch size: $4096\ \textnormal{effective} = 32 \times 2\ \textnormal{GPUs} \times 64\ \textnormal{accumulation steps}$
    \item Learning rate (text encoder): $7.5\times 10^{-4}$
    \item Learning rate (other encoders): $10^{-3}$
    \item Precision: bf16 mixed
    \item Seed: 5501650
    \item Adam optimizer:\vspace{-0.3em}
        \begin{itemize}
        \itemsep0em 
            \item Epsilon: $10^{-8}$
            \item Weight decay: $0.1$
            \item Betas: [0.9, 0.999]
        \end{itemize}
\end{itemize}

We use two distinct learning rates because a lower value is commonly recommended for text-only pretraining \citep{roberta,bert}, while a higher one was originally used for multimodally pretraining FLAVA. Multiple strategies for correctly choosing modality-specific learning rates are treated extensively in \citet{one_lr_per_modality}, where the \textit{"Keep" Strategy} (ours) is among the most straightforward of them. While simple, it outperforms (in the authors' empirical study) the global learning rate strategy as it ensures that each unimodal subpart still has effective gradients when training the fusion model.

\def\software#1{\texttt{#1}}

\paragraph{Software} We use \software{Pytorch Lightning} \citep{Falcon_PyTorch_Lightning_2019} as the main training framework and \software{Weights and Biases} \citep{biewald2020experiment} to track relevant metrics in real time. We use the Huggingface \software{datasets} library \citep{Lhoest_Datasets_A_Community_2021} to interleave the modality-specific datasets, and the HuggingFace \software{transformers} library \citep{Wolf_Transformers_2020} to access and train randomly-initialized FLAVA models.  

\paragraph{Hardware} 
We run each training job in Distributed Data-Parallel mode, across two NVIDIA Tesla A100 Ampere 40 GB graphics cards, on the same node, in ETH Zürich's Euler datacenter. 
For each of the two GPUs, there are 4 CPU workers loading data (this number was empirically found to be optimal), with each CPU worker having 10GB of RAM available. 
The average runtime for our jobs running on 100M words was six days, and for 10M words, it was three days. 
Thus, we count a total of (2 GPUs) * (8 jobs) * (108 hours / job) = 1728 GPU hours used to train the models reported in this study, not counting our hyperparameter search.

\paragraph{Dataloader Sampling Weights} 
During multimodal pretraining, we alternate samples from three data loaders with independent weights, initialized (and normalized) proportional to their sizes. 
For example, for the condition with 100M words and 40K images (hence, all images and 10M words of text are paired), we would have the following initial sampling weights: 0.833 (text), 0.083 (vision), 0.083 (multimodal).
For maximal text encoder performance, we perform a hyperparameter search and determine a simple rule-based approach to further improve the distribution of the sampling weights: 
If text is not the predominant modality, we change it to the uniform distribution; otherwise, we leave the initial weights unchanged.

\begin{table}[t]
    \centering
    \begin{tabular}{lllll}
    \toprule
         \diagbox[width=6em]{Words}{Images} & None & 40K & 400K & 4M \\ \midrule
        10M & 4604  & 5346 & 4876 & 4876 \\ 
        100M & 23818 & 12267 & 16344 & 16542 \\\bottomrule
    \end{tabular}
    \caption{Model checkpoints (training step \#) chosen for evaluation based on the masked language modeling validation loss (Figure \ref{fig:validation_losses}).}
    \label{tab:best_checkpoints}
\end{table}

\begin{figure}[t!]
    
    \minipage{0.5\textwidth}
        \includegraphics[width=\linewidth]{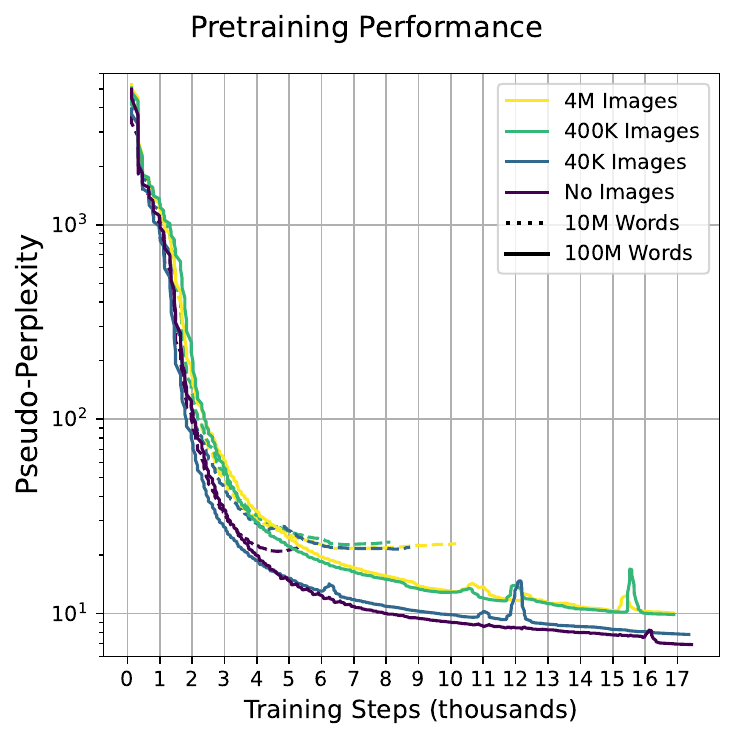}
    \endminipage\hfill
    \caption{PPPL performance for the two data volumes of 10M and 100M words. The training steps on the x-axis are counted across all objectives.}
    \label{fig:pppl_eval}
    \minipage{0.5\textwidth}
        \includegraphics[width=\linewidth]{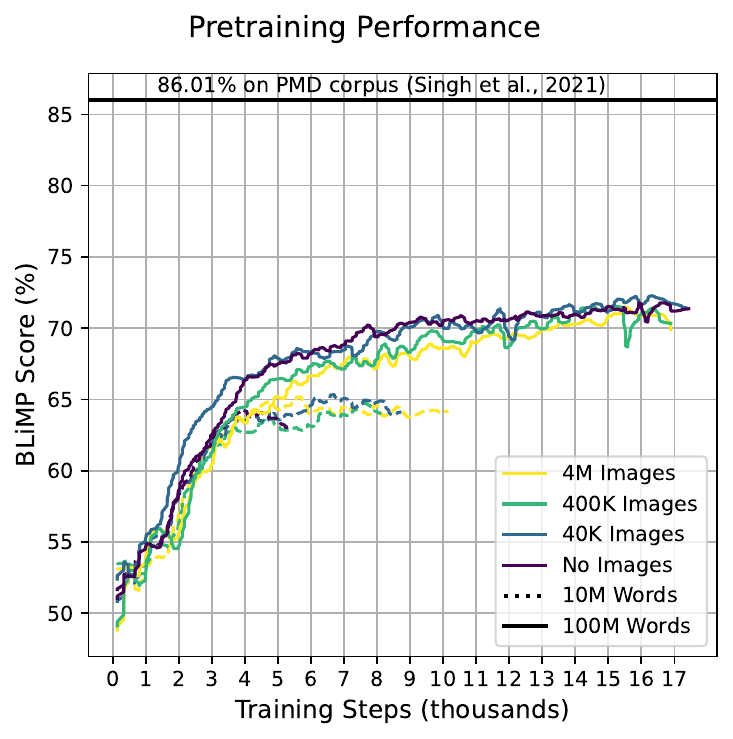}
    \endminipage\hfill
    \caption{BLiMP performance for the two data volumes of 10M and 100M words. The training steps on the x-axis are counted across all objectives.}
    \label{fig:blimp_eval}
\end{figure}

\paragraph{Modality-Specific Early Stopping}
We develop custom logic to prevent the models from overfitting on any given modality. 
For example, when the sampling rate is 0.083 for the multimodal and text data loaders yet 0.833 for the vision data loader, the former two modalities will likely begin to overfit well before the latter. 
To avoid this, we detect increases in validation loss and (each time) halve the corresponding task's sampling weight. 
If the validation loss continues to steadily increase after three validation steps, we set the task weight to 0. 
To prevent catastrophic forgetting of the multimodal input, we allow the models to restart training on vision and multimodal data after a certain period of inactivity (here, 10 validation phases).

\paragraph{Model Selection} 

For each of the eight input configurations, we select the best model checkpoints for evaluation based on the lowest recorded masked language modeling loss on the validation set (see Figure \ref{fig:validation_losses} for validation losses for every training objective and model). 
Table \ref{tab:best_checkpoints} shows the number of training steps for the selected checkpoints from each configuration. For additional information, we also regularly evaluate the models' pseudo-perplexity on a held-out test set (see Figure \ref{fig:pppl_eval}).

\section{Results}

\begin{table*}[!t]
    \centering
    \resizebox{\textwidth}{!}{%
    \begin{tabular}{ccccccccc}
    \toprule
         \multirow{2}{*}{\centering \textbf{Tasks}} & \multicolumn{4}{c}{\textbf{10M Words}} & \multicolumn{4}{c}{\textbf{100M Words}} \\
          & \textbf{None} & \textbf{40K} & \textbf{400K} & \textbf{40M} & \textbf{None} & \textbf{40K} & \textbf{400K} & \textbf{40M} \\ \cmidrule(lr){1-1}\cmidrule(lr){2-5}\cmidrule(lr){6-9}
        (Super)GLUE (Acc., F1, MCC) & 65.49 & 64.68 & 65.17 & \underline{65.76} & \underline{70.42} & 69.24 & 68.9 & 69.07 \\
        BLiMP (Acc.) & 63.96 & 63.98 & 63.31 & \underline{64.53} & 71.32 & 70.45 & \underline{71.9} & 70.93 \\
        MSGS (MCC) & -12.88 & -12.16 & \underline{-8.84} & -18.62 & -8.66 & \underline{-6.18} & -7.41 & -7.47 \\
    \bottomrule
\end{tabular}}\caption{Performance for each of the eight models on the BabyLM test suites (detailed version in Table \ref{tab:finetuning_detailed}).}
    \label{tab:highlev_finetuning}
\end{table*}

We evaluate the selected checkpoints from all eight training configurations on the BabyLM evaluation pipeline \citep{warstadt2023findings}, including evaluations on benchmarks for grammar \citep[BLiMP;][]{warstadt2020blimp}, language understanding \citep[GLUE and SuperGLUE;][]{wang2019glue,wang2019superglue}, and linguistic generalization \citep[MSGS;][]{warstadt2020learning}. 
For BLiMP and pseudo-perplexity \citep{wang2019glue}, we also report intermediate results for all of the training checkpoints.

Our overall results in Table \ref{tab:highlev_finetuning} largely confirm earlier work finding that vision is, at best, not consistently helpful to language performance. 
With a data volume of 10M words, FLAVA does sometimes perform marginally better on grammar-oriented tasks in the presence of visual cues. 
For other evaluations and with a data volume of 100M words, we also find no consistent advantages in our experimental setting. 
Of those improvements we do observe, our tests deem it unlikely that they are due to cross-situational learning (see Section \ref{sec:grounding}).

\begin{figure*}[h!]
\minipage{\textwidth}
\includegraphics[width=\linewidth]{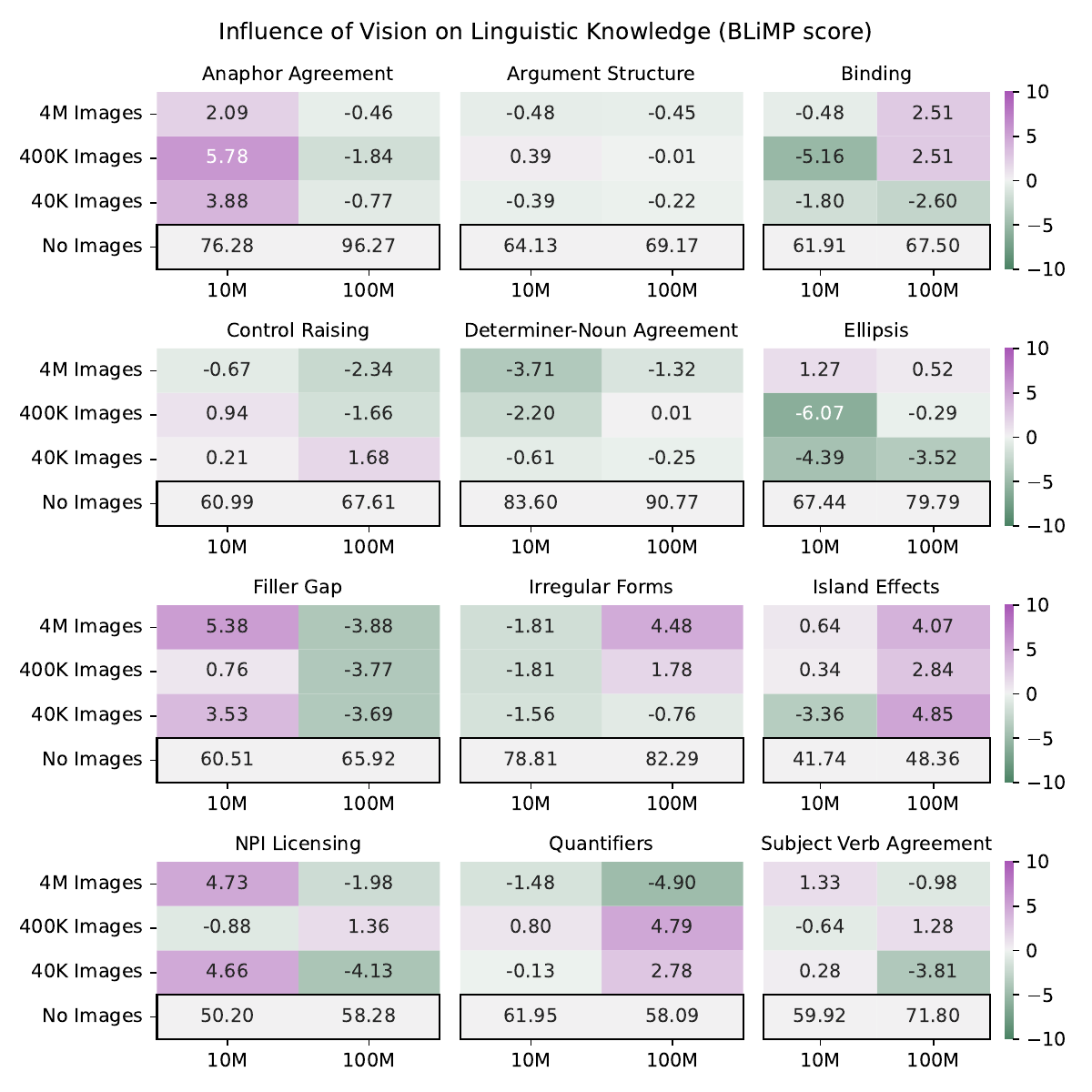}
\caption{Zero-shot accuracies, in percentages, obtained on the BLiMP task for each grammatical category (x12) and FLAVA run configuration of input text volume (10M and 100M words) and input vision volume (0, 40K, 400K and 4M images). The model checkpoints used to generate these results were selected as described in Table \ref{tab:best_checkpoints}.}
\label{fig:detailed_blimp}
\endminipage
\end{figure*}
\subsection{Pseudo-perplexity} \label{subsec:pppl}

For validation, Figure \ref{fig:pppl_eval} shows the pseudo-perplexity \cite[PPPL;][]{pll_original} per token on a held-out evaluation subset of WIT throughout training. 
Unsurprisingly, PPPL is lower (better) for the 100M word models compared to the 10M word models. 
Additionally, the metric appears to converge for the 10M word models, while it may still be decreasing for 100M word models.%
\footnote{Our scheduler triggered early stopping based on validation loss despite the apparent possibility that longer training might have been beneficial. 
More generally, there are many potential improvements to task scheduling and early stopping for multitask learning that we leave to future work.} 
The most unexpected finding is that PPPL is consistently worse as the amount of image data increases for a given amount of text data throughout training.
This degradation may suggest that our multitask training procedure causes the models to sacrifice MLM performance in favor of other objectives as the proportion of visual and multimodal samples increases.

\subsection{Grammaticality}

We evaluate linguistic knowledge using BLiMP \citep{warstadt2020blimp}, which tests the ability of models to distinguish grammatical sequences from minimally different ungrammatical ones in a zero-shot setting.  

Table \ref{tab:highlev_finetuning} shows the overall BLiMP performance from each condition.
We notice that text quantity makes a big difference in performance. 
Changes in vision, on the other hand, are associated with small amounts of variation that are sometimes positive or negative. 
Hence, due to the lack of a consistent pattern and the small number of runs, we cannot confidently conclude that vision causes an increase or decrease in performance. 

Figure \ref{fig:blimp_eval} shows the BLiMP results for each validation step throughout training. 
For most of the duration of training, particularly for the 100M word models, models with less image data perform better.
This behavior mostly matches the pattern we observe for pseudo-perplexity, except that the differences seemingly disappear by the end of training. 
This result confirms earlier findings that \mbox{(pseudo-)perplexity} is not entirely predictive of grammatical knowledge \citep{hu2020systematic}.

Individual BLiMP categories are more closely compared in Figure \ref{fig:detailed_blimp}.
Previous work shows that phenomena related to agreement have the steepest learning curves at the 10M word scale \citep{zhang2020you}.
Therefore, if the hypothesis that vision accelerates LM learning is correct, we might expect to see the greatest signs of improvement for 10M word models on this subset of test suites.
Figure \ref{fig:detailed_blimp}, however, shows conflicting and inconclusive results, with improvements in \textit{anaphor agreement} but a slight degradation for \textit{determiner-noun agreement}, and little change for \textit{subject-verb agreement}. 

We observe that multimodal pretraining may have a regularizing effect at smaller data scales: 
BLiMP performance improves at times although the pseudo-perplexity (i.e., test loss) is consistently higher (by 1-3 units) for the vision-infused models. 
Moreover, the vision-infused models run for almost twice as long before starting to overfit (8k v.s. 4k steps), gaining accuracy in areas such as \textit{anaphor agreement}, \textit{filler gap dependencies}, and \textit{NPI licensing}, although not so on test suites such as \textit{argument structure} and \textit{subject-verb agreement}.

\subsection{Fine-Tuning Evaluations}
In addition to the above, we also use GLUE/SuperGLUE \citep{wang2019glue,wang2019superglue} and MSGS \citep{warstadt2020learning} to fine-tune and evaluate all eight models on a selection of downstream tasks that focus on language understanding and linguistic generalization.

As expected, the results in Table \ref{tab:highlev_finetuning} show that overall GLUE performance increases (by around 5\%) at higher text data scales. Within each of the two text volume groups, however, there is no reliable improvement due to the addition of vision, though vision-infused models appear to be slightly better (relatively) at lower data scales, of up to 10M words. Generally, the models perform similarly on the selected downstream tasks (performance after fine-tuning), in line with BLiMP results. 

Scores on MSGS are negative for all models, for all ambiguous subtasks (i.e., those subtasks not in the control condition), as shown in Table \ref{tab:finetuning_detailed}. This indicates that all of our models are consistently biased towards generalizing based on shallow surface cues rather than linguistic features.

\subsection{Cross-Situational Learning} \label{sec:grounding}

To assess the \textit{symbolic grounding} of our models, for every input configuration checkpoint in Table \ref{tab:best_checkpoints}, we evaluate the multimodal text retrieval zero-shot accuracy on ImageNet-1k \citep{russakovsky2015imagenet}. The goal is to select, for every given \textit{query} image, the best-fitting text caption from a pool of 1000 options. To this end, we compute cosine similarities as matching scores between the queried image's representation and the representations of 1000 template-averaged\footnote{Since the captions for even a specific entity can vary, e.g., \textit{a doodle of a car}, \textit{a photo of a large car}, etc. we compute an average over $\sim$80 such templates for each entity.} potential captions. Lastly, we retrieve the text caption with the highest matching score for each image query. We follow \citet{radford2021learning} to calculate the zero-shot accuracy.

As a baseline, we assess FLAVA pretrained on the PMD corpus and obtain \textit{top1} and \textit{top5} accuracies of 32\% and 60\%, respectively. The models we pretrain, however, obtain average \textit{top1} and \textit{top5} accuracies of 0.1\% and 0.5\%, respectively. Some of the possible factors responsible for this random guess performance could be: 1) the multitask scheduler described in Appendix \ref{subsec:training_config} was misconfigured (this aligns with findings in \citet{msc_thesis}), 2) the smaller magnitude of the training data (WiT is a subset of PMD), 3) the weak alignment between some of the text (full paragraphs) and the corresponding images (Imagenet-1k only evaluates caption alignments), or 4) the fact that we do not pretrain the vision encoder unimodally on ImageNet-1k, as in \citet{flava}. 

\section{Conclusion} 
We perform an ablation study on a state-of-the-art, multimodal language model under varying text and vision configurations. Our training recipe avoids the problem of catastrophic forgetting of complex language, which previous approaches fell prey to, by performing multitask training on both multimodal and unimodal tasks in a more diverse domain. Nonetheless, our results largely confirm earlier work finding that vision is (at best) not consistently helpful to language performance. During pretraining at the small 10M word scale, the FLAVA architecture \citep{flava} does sometimes appear to perform marginally better on grammar-oriented tasks in the presence of visual cues. However, for other evaluations and with a data volume of 100M words, we find no consistent advantages in our experimental setting. 

At the small data scales that we pretrain our models in this study (up to 100M words and the corresponding images), our tests in Section \ref{sec:grounding} deem it unlikely that the models are benefiting from cross-situational learning. Alternatively, the extra parameters in the multimodal encoder could simply be increasing FLAVA's modeling capacity, a hypothesis that we leave for future work. Regardless, multimodal pretraining seems to exhibit a regularizing effect: although pseudo-perplexity is consistently worse for the vision-infused models, grammatical performance fluctuates and is often at least as good.

We conclude that the lack of visual input alone does little to explain the large data efficiency gap between LMs and humans observed in grammar learning, though we leave open the possibility that this conclusion will change with better architectures and techniques for integrating vision and language at training time. 

\section*{Limitations} \label{sec:limitations} 

The robustness of the observations made in this report is limited by the fact that each configuration (text/vision input volume) was only run once. Future work should provide at least 5 re-runs per configuration (with different seeds), as there can be considerable variance even in different models with the same configuration \citep{mccoy-etal-2020-berts}. Due to the computational intensity of performing re-runs, this was not possible in time for this submission. 

Significant GPU resources are required to effectively train large language models, partly because of the large batch sizes and the scale of the datasets. In this work, we use $\approx$ 1728 GPU hours on very recent hardware (further details in Section \ref{subsec:training_config}).

Finally, there is an architectural difference between the unimodal and multimodal models in our experiments. The unimodal models are trained entirely without the visual or multimodal encoders. Although these parameters are not used by the multimodal model during evaluation on language-only tasks, they are used during training, and so they may have an indirect effect on what the language encoder learns. To test whether the potential performance improvements in grammaticality and language understanding can indeed be attributed to the visual cues or rather simply to the increased number of parameters in the multimodal encoder, future work should conduct additional baseline experiments, e.g., where the images are replaced with random noise pixels.


\bibliographystyle{acl_natbib}
\bibliography{anthology,custom}

\appendix



\clearpage
\section{Pretraining Validation Losses} \label{apx:val_losses}

\begin{figure}[!htb]
\minipage{\textwidth}
\includegraphics[width=\linewidth]{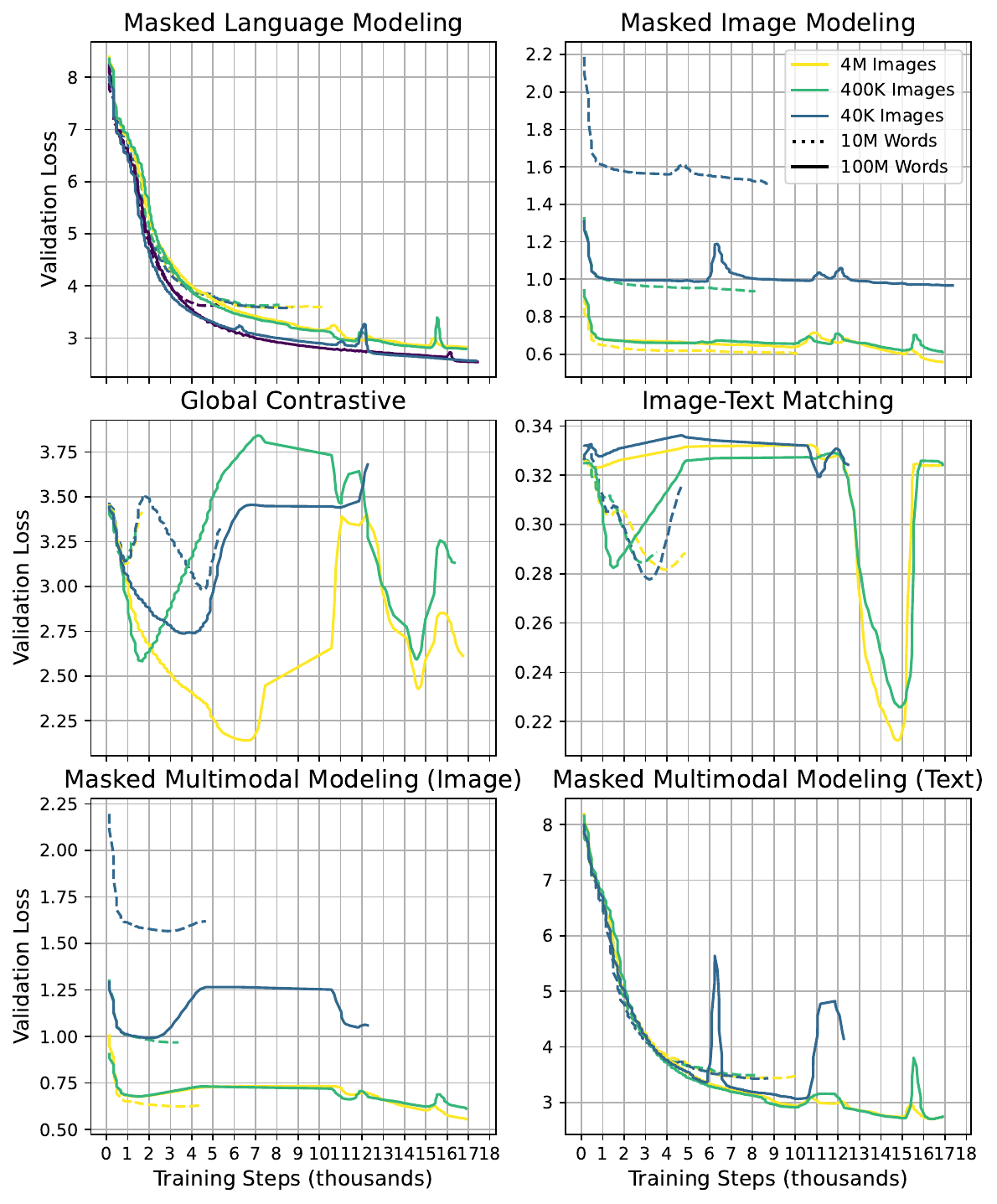}
\caption{Validation losses for every training objective on a held-out set. While the MLM -- and to a certain extent, also the MMM (Text) -- losses are closely proportional to the pseudo-perplexity metric in Figure \ref{fig:pppl_eval} (including some occasional spikes associated with checkpoint loading), the other losses are less stable. We point out some issues with the scheduler mechanism in Sections \ref{subsec:pppl} and \ref{sec:grounding}.}
\label{fig:validation_losses}
\endminipage
\end{figure}

\clearpage
\section{Fine-tuning Performance} \label{apx:fine_tuning}


\begin{minipage}{\textwidth}
\centering
    \resizebox{\textwidth}{!}{%
        \begin{tabular}{llcccccccc}
        \toprule
             \multirow{2}{*}{\centering \textbf{Task}} & \multirow{2}{*}{\centering \textbf{Subtask}} & \multicolumn{4}{c}{\textbf{10M Words}} & \multicolumn{4}{c}{\textbf{100M Words}} \\ 
             & & \textbf{None} & \textbf{40K} & \textbf{400K} & \textbf{40M} & \textbf{None} & \textbf{40K} & \textbf{400K} & \textbf{40M} \\ \cmidrule(lr){1-2}\cmidrule(lr){3-6}\cmidrule(lr){7-10}
             
        \multirow{11}{*}{\rotatebox{90}{(Super)GLUE}}
            &BoolQ (accuracy) & 64.32 & 65.70 & 65.98 & \underline{66.8} & \underline{69.43} & 67.08 & 65.70 & 66.25 \\
            &CoLA (MCC) & \underline{4.21} & -4.16 & 0.00 & 0.00 & \underline{28.43} & 21.56 & 20.00 & 20.93 \\
            &MNLI (accuracy) & \underline{73.27} & 72.46 & 70.07 & 71.98 & \underline{74.63} & 71.85 & 74.57 & 74.11 \\
            &MNLI-mm (accuracy) & 72.5 & 73.71 & \underline{73.98} & 73.74 & \underline{78.06} & 77.83 & 76.78 & 75.84 \\
            &MRPC (F1) & 81.61 & \underline{83.10} & 81.61 & 82.00 & 82.99 & \underline{85.51} & 84.78 & 84.98 \\
            &MultiRC (accuracy) & 59.58 & 60.35 & 61.45 & \underline{62.32} & \underline{67.47} & 62.76 & 64.62 & 67.25 \\
            &QNLI (accuracy) & \underline{80.88} & 79.40 & 80.31 & 79.97 & \underline{83.90} & 82.76 & 82.55 & 81.98 \\
            &QQP (F1) & 81.76 & 82.28 & 82.49 & \underline{82.57} & 82.99 & \underline{84.30} & 82.91 & 82.43 \\
            &RTE (accuracy) & \underline{57.58} & 53.54 & 53.54 & 55.56 & 53.54 & 57.58 & \underline{58.59} & 56.57 \\
            &SST-2 (accuracy) & 83.27 & 83.66 & 84.84 & \underline{87.01} & \underline{91.73} & 88.98 & 87.20 & 87.99 \\
            &WSC (accuracy) & 61.45 & 61.45 & \underline{62.65} & 61.45 & \underline{61.45} & \underline{61.45} & 60.24 & \underline{61.45} \\ \cmidrule(lr){1-2}

        \multirow{12}{*}{\rotatebox{90}{BLiMP (Acc.)}}
            & Anaphor Agreement & 76.84 & \underline{79.55} & 76.99 & 77.04 & 94.89 & 95.35 & 91.10 & \underline{96.27} \\
            & Argument Structure & \underline{64.14} & 62.38 & 62.55 & 62.45 & 69.58 & \underline{69.77} & 68.15 & 68.38 \\
            &Binding & 62.07 & 63.79 & \underline{64.00} & 60.98 & \underline{68.91} & 66.62 & 65.67 & 67.99 \\
            &Control/Raising & 61.4 & 62.06 & \underline{64.54} & 63.10 & 67.96 & \underline{69.64} & 67.63 & 67.87 \\
            &Determiner Noun Agreement & 82.91 & 80.42 & 82.11 & \underline{83.00} & 90.27 & \underline{91.94} & 88.45 & 89.49 \\
            &Ellipsis & \underline{68.48} & 65.76 & 61.61 & 66.22 & \underline{81.12} & 78.12 & 80.60 & 77.37 \\
            &Filler Gap Dependencies & 59.52 & 59.73 & 57.91 & \underline{60.85} & \underline{67.24} & 65.41 & 64.19 & 61.92 \\
            &Irregular Forms & \underline{83.51} & 75.98 & 78.73 & 74.45 & 84.17 & 84.12 & 83.66 & \underline{84.89} \\
            &Island Effects & 40.43 & 44.13 & 40.28 & \underline{48.02} & 51.83 & \underline{53.10} & 48.47 & 48.65 \\
            &NPI Licensing & 52.49 & 54.43 & \underline{54.66} & 49.53 & 61.24 & \underline{64.79} & 60.04 & 54.68 \\
            &Quantifiers & 58.55 & 57.86 & \underline{64.76} & 57.73 & \underline{60.10} & 58.40 & 57.16 & 58.50 \\
            &Subject Verb Agreement & 60.29 & 61.72 & \underline{61.81} & 61.37 & \underline{75.34} & 72.65 & 70.19 & 69.18 \\ \cmidrule(lr){1-2}

        \multirow{9}{*}{\rotatebox{90}{MSGS (MCC)}}
            &Control Raising (control) & 21.90 & 27.36 & \underline{31.42} & 22.24 & 46.21 & 45.25 & 46.85 & \underline{55.00} \\
            &Control Raising--Lexical Content & -46.54 & -18.71 & \underline{-13.17} & -72.82 & \underline{-61.04} & -66.26 & -63.65 & -88.49 \\
            &Contro Raising--Relative Position & -98.69 & \underline{-97.82} & -98.33 & -100.00 & -99.90 & \underline{-89.06} & -97.74 & -99.34 \\
            &Lexical Content (control) & 100.00 & 100.00 & 100.00 & 100.00 & 100.00 & 100.00 & 100.00 & 100.00 \\
            &Main Verb (control) & 86.84 & \underline{96.84} & 93.24 & 86.92 & \underline{99.85} & 99.40 & 96.54 & 98.08 \\
            &Main Verb--Lexical Content & -100.00 & -100.00 & -100.00 & -100.00 & -100.00 & -100.00 & -100.00 & -100.00 \\
            &Main Verb--Relative Position & \underline{-86.73} & -88.78 & -91.56 & -95.32 & -98.41 & \underline{-86.56} & -95.64 & -93.18 \\
            &Relative Position (control) & \underline{96.78} & 81.47 & 90.28 & 89.14 & 99.98 & 99.47 & \underline{100.00} & 99.98 \\
            &Syntactic Category (control) & \underline{57.04} & 13.15 & 38.77 & 28.12 & 72.40 & 73.32 & 77.37 & \underline{89.29} \\
            &Syntactic Category--Lexical Content & -100.00 & -81.37 & \underline{-77.31} & -95.49 & -81.08 & \underline{-74.66} & -79.96 & -78.53 \\
            &Syntactic Category--Relative Position & -72.34 & \underline{-65.91} & -70.53 & -67.61 & -73.29 & -68.86 & -65.26 & \underline{-64.96} \\
        \bottomrule
        \end{tabular}
    }
    \captionof{table}{Detailed fine-tuning performance for each of the eight models. F1 denotes macro-F1, MCC denotes Matthew's correlation coefficient, and random chance accuracy on all BLiMP tasks (i.e., the second group) is 50.}
    \label{tab:finetuning_detailed}
\end{minipage}

\end{document}